# Shallow Encoder Deep Decoder (SEDD) Networks for Image Encryption and Decryption

Chirag Gupta
chirag@cgupta.tech

**Abstract**

This paper explores a new framework for lossy image encryption and decryption using a simple shallow encoder neural network E for encryption, and a complex deep decoder neural network D for decryption. E is kept simple so that encoding can be done on low power and portable devices and can in principle be any nonlinear function which outputs an encoded vector. D is trained to decode the encodings using the dataset of image - encoded vector pairs obtained from E and happens independently of E. As the encodings come from E which while being a simple neural network, still has thousands of random parameters and therefore the encodings would be practically impossible to crack without D. This approach differs from autoencoders as D is trained completely independently of E, although the structure may seem similar. Therefore, this paper also explores empirically if a deep neural network can learn to reconstruct the original data in any useful form given the output of a neural network or any other nonlinear function, which can have very useful applications in Cryptanalysis. Experiments demonstrate the potential of the framework through qualitative and quantitative evaluation of the decoded images from D along with some limitations.

## 1. Introduction

Cryptography is concerned with encoding a sensitive piece of data in a form which is unintelligible or meaningless to any human or machine other than the intended party which has the decoding mechanism to regenerate the original data from the encoding. Cryptanalysis on the other hand is used to breach cryptographic security systems and gain access to the contents of encrypted messages [1]. A primary research area in Cryptography and Cryptanalysis is encryption and decryption techniques for secure communication and code breaking. Most of the protocols for encryption in commercial use today are n bit key based or hash functions like RSA, DES, 3DES, AES, SHA. In this paper we are concerned with images as data to be encrypted at source, sent securely and decrypted at destination [2].

Deep learning neural networks have been an area of active research as a means for both encryption and decryption for some time. Techniques like Hopfield neural networks, chaotic time delayed neural networks [3], Autoencoder networks [4], [5], Generative and GAN based [5], [6] etc have shown encouraging results. Autoencoders and generative networks are specially interesting as the proposed framework is inspired from them. However, this research has not found much application outside research where conventional algorithms continue to dominate. One of the reasons for this is that deep neural networks as in autoencoders are computationally expensive to run involving many matrix operations [7]. This is especially true for discrete mobile security systems such as miniature cameras used in espionage, cameras used in drones or other robotic systems, portable communication devices in military, space as well as general real-time applications. The proposed framework attempts to solve this problem. For the purpose of breaking encryption, deep learning models are especially suited because of their inherent ability to learn nonlinear functions and abstract relations given a large number of labelled examples [8].

In the proposed *Shallow Encoder – Deep Decoder (SEDD)* framework, the encoder E is made extremely simple and shallow to reduce the computational load for encryption considerably and can be done on low powered mobile ARM processors in real time. The weights and biases of E are randomly initialized and not trained to optimize any objective, as the task of E is to process the $n$ dimensional image vector into a $p$ dimensional floating-point encoding vector which represents the output layer of E. It is not possible to reverse engineer the image from the encoding without knowing the parameters and the overall structure of E. Further brute force, statistical and differential attacks are impractical as the network has piecewise linear units with nonlinear activations and thousands of randomly initialized parameters [9]. The decoder D is a generative deep neural network which takes an encoded vector as input and outputs an $n$ dimensional image vector which is reshaped into an 8-bit RGB matrix and processed into an image. D is trained as a regression problem on a large number of image-encoding pairs obtained from E to generate the original image from the encoding. Therefore, D is not dependent on E for training in any way, provided a large dataset of encodings from E is available and can be thought of as an adversary net trying to learn the underlying hidden relations in E which convert the image to the encoding [10].

## 2. Related Work

A lot of research has gone into deep learning models for Cryptography and Cryptanalysis applications in recent years. However, much of the research ties conventional encryption techniques with neural networks where the latter serve as a booster or enhance the technique instead of the full focus being on deep learning methods.

Autoencoders have been developed for encoding and decoding data using neural networks [11]. Applications of autoencoders in speech such as speech spectrogram coding [12], for generation images [13] and for denoising have given useful results [14], [15]. Stacked autoencoders have also been used for encryption but are computationally expensive for encoding which limits their application and are less secure as the encoder and decoder are trained in tandem. However, these autoencoders are mostly lossless with decrypted images having high quality [5]. But applications of autoencoders in cryptography in the way proposed by this paper remains mostly unexplored.

Other deep learning approaches to encryption such as chaotic Hopfield neural networks generate binary sequences to mask plain text [3]. An older but relevant 'Analysis of Neural Cryptography' [16] is based on mutually learning networks but is prone to attacks.

Aside from cryptography, generative networks for images such as Plug & Play generative networks [17], GANs (generative adversarial networks) [10], [18] can generate photorealistic images with random noise as input. SEDD networks build upon these generative networks, without having the discriminator but trading off quality of the generated images.

## 3. Encoder

The SEDD framework consists of a shallow Encoder network E which is a feedforward single hidden layer perceptron. Therefore the 3 layers in the encoder are $E_I$ (input layer), $E_H$ (hidden layer), $E_O$ (output layer). $E_I$ takes an RGB 8-bit image flattened into a single $n$ dimensional vector. $E_H$ is a hidden layer with a small size to reduce the number of parameters and hence the computations required to get the output. $E_O$ is the output layer with size equal to the desired encoding size $p$. Choosing a large $p$ increases the complexity but also increases the features available to the decoder for training. The complexity of E is kept low as it is to run on device in real time. Weights and biases of E are randomly initialized, and the model (E) is saved as such without any training. Therefore, E serves as a function which is highly nonlinear and mangles the input to encrypt it. The encodings from E are considered as encrypted data and it is not possible to recreate images back from encodings without the decoder.

## 4. Decoder

The Decoder D is the main workhorse of the SEDD framework and is a deep multi-layer neural network (here for simplicity D is a multi-layer perceptron, more complex networks are discussed in Future Work). The input layer of D, $D_I$ is $p$ dimensional as it takes the encoding vector as input. D is trained on image-encoding pairs available from E and tries to recreate images from the encodings and therefore the output contains $n$ units which is reshaped into an RGB 8-bit image. D runs on a machine with high computational power and is saved in a secure way with the intended agent.

## 5. Algorithm

The encoder E contains 3 layers: input layer $E_I$, hidden layer $E_H$ and output layer $E_O$. Let the 8-bit RGB image $\mathbf{A} \in \mathbb{R}^{h \times w \times 3}$ of height $h$ and width $w$ is to be encoded. Therefore, the flattened image vector is $n$ dimensional which is the same as the size of $E_I$, where $n = h \times w \times 3$. Let this vector be $\mathbf{a}$.

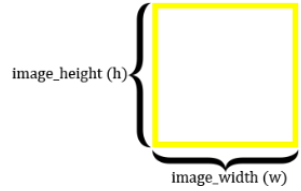

**Figure 1:** Sample image in dataset

$$\Rightarrow \begin{bmatrix} [A_{(1,1)r}\ A_{(1,1)g}\ A_{(1,1)b}] & \cdots & [A_{(1,w)r}\ A_{(1,w)g}\ A_{(1,w)b}] \\ \vdots & \ddots & \vdots \\ [A_{(h,1)r}\ A_{(h,1)g}\ A_{(h,1)b}] & \cdots & [A_{(h,w)r}\ A_{(h,w)g}\ A_{(h,w)b}] \end{bmatrix} \Rightarrow flatten(A) \Rightarrow \mathbf{a} = \begin{bmatrix} a_1 \\ \vdots \\ a_n \end{bmatrix} \quad (1)$$

$E_H$ has a size of $h$ which is kept small to make E computationally simple. A typical value of $h$ will be 10. The size of output layer $E_O$ is the desired size of the encoding vector $\mathbf{x}$. E's weights and biases are initialized randomly and saved right away. $\mathbf{x}$ is called the encoding or encrypted vector i.e. the image has been converted into a vector of floating-point numbers represented by $\mathbf{x}$.

$$\mathbf{a} = \begin{bmatrix} a_1 \\ \vdots \\ a_n \end{bmatrix} \Rightarrow E \left\langle E_I \left| \begin{array}{c} E_H,\ units = h \\ ------- \\ Output = Relu(\mathbf{a}) \end{array} \right| \begin{array}{c} E_O,\ units = p \\ ----------------- \\ Output, \mathbf{x} = Sigmoid(Output(\ E_H)) \end{array} \right\rangle \Rightarrow \mathbf{x} = \begin{bmatrix} x_1 \\ x_2 \\ \vdots \\ x_p \end{bmatrix}, \mathbf{x} \in \mathbb{R}^p \quad (2)$$

The job of decryption or decoding the encoded vector back into the original image lies with the decoder D. Therefore D is a complex deep neural network with $k$ hidden layers $D_{H1}, \ldots, D_{Hk}$. D takes the encoded vector $\mathbf{x}$ as input and therefore $D_I$ is $p$ dimensional. As D is a generative network (it outputs a vector which is reshaped into an image), the hidden layers are activated by the leaky rectified linear function as it tends to give better results in such networks [18]. The hidden layers are regularized with adding dropout layers after activations [19]. The output layer $D_O$ is $n$ dimensional, where $n = h \times w \times 3$ and the output vector is reshaped back into a matrix of an RGB 8-bit $h \times w$ image.

D is trained on a dataset $\mathbf{X} \in \mathbb{R}^{m \times p}$ of $m$ image-encoding pairs obtained from E. If $\mathbf{A}_{images}$ is the set of images,

$$A_{images} = \begin{bmatrix} A_1 \\ A_2 \\ \vdots \\ A_m \end{bmatrix} \Rightarrow Encoder(A_{images}) \Rightarrow X = \begin{bmatrix} x_1 \\ x_2 \\ \vdots \\ x_m \end{bmatrix}, X \in \mathbb{R}^{m \times p} \quad (3)$$

As D is trained on the dataset **X**, for the $i^{th}$ forward pass for the image $A_i$, $x_i$ is the input image into $D_I$ and $y_i$ is the output vector from $D_O$. $y_i$ is reshaped into the image matrix $\hat{A}_i$. Stochastic gradient descent (SGD) is used to minimize the loss which is the mean squared error (MSE) and is calculated by comparing the $y_i$ (or $\hat{A}_i$) to $A_i$. The training is done till a global minimum of loss is reached for the test set. D would be ideal and lossless if $\hat{A}_i$ is equal to $A_i$ i.e. we are able to extract the exact image from the encoding.

$$x_i^T = [x_{i1} \quad x_{i2} \quad \cdots \quad x_{in}] \quad (4)$$

$$\Rightarrow D \left( D_I \left| D_{H1} \quad \cdots \quad \underbrace{\overline{LeakyRelu(W_{Hj}^T Output(D_{H(j-1)}) + b_{Hj})}}_{Dropout}^{D_{Hj}, \ units = h_j} \quad \cdots \quad D_{Hk} \left| \overline{y = Sigmoid(W_O^T Output(D_{Hk}) + b_O)}^{D_O, \ units = n} \right. \right. \right) \quad (5)$$

$$\Rightarrow y_i^T = [y_{i1} \quad y_{i2} \quad \cdots \quad y_{in}] \Rightarrow reshape(y_i \to h \times w \times 3) \quad (6)$$

$$\Rightarrow \hat{A}_i = \begin{bmatrix} [\hat{A}_{(1,1)r} \ \hat{A}_{(1,1)g} \ \hat{A}_{(1,1)b}] & \cdots & [\hat{A}_{(1,w)r} \ \hat{A}_{(1,w)g} \ \hat{A}_{(1,w)b}] \\ \vdots & \ddots & \vdots \\ [\hat{A}_{(h,1)r} \ \hat{A}_{(h,1)g} \ \hat{A}_{(h,1)b}] & \cdots & [\hat{A}_{(h,w)r} \ \hat{A}_{(h,w)g} \ \hat{A}_{(h,w)b}] \end{bmatrix} \quad (7)$$

E runs on the edge (on device or companion device on which the image is created or received). Once the encoding is obtained the original image can be deleted. This algorithm is designed for the encryption to run on low power portable devices such as an ARM processor in a miniature camera or a raspberry pi like computer. The encodings can then be transferred safely to the intended party which has the trained decoder. The decoded images can be obtained from D by doing a forward pass (inference) in the decoder.

## 6. Experiments

Tensorflow 1.14 with keras on python 3.6.9 was used to implement the framework which was trained on an nVidia GTX 1660ti GPU. The decoder was trained on a range of datasets including MNIST [20], CIFAR-10 [21] and Cat image dataset [22]. The implementation here is shown for the cat images. The Cat image dataset contains 12500 images of cats for training.

The encoder E is randomly initialized with the hidden layer $E_H$ having 10 units i.e. $h = 10$ with *relu* activation. The output layer of E has sigmoid activation with the size of 1024. Hence the encoding size $n$ is 1024. There are just 15,774 parameters in E. The images in the cat dataset are resized into a 150 × 150 matrix with h = w = 150. The decoder D has 3 hidden layers with 512 units each with leaky rectified linear activation having an alpha of 0.2 and a Dropout layer with a rate of 0.3, 0.3 and 0.2 respectively. The loss mse is minimized during training of D using SGD. There are 36,727,212 trainable parameters in D.

Training for a large number of epochs (>20) quickly overfits artificially lowering the train mse a lot so an Early stopping with a *test mse* < 0.075 constraint is applied (this value is derived empirically). Under this constraint, D trains for 10 epochs and achieves a satisfactory loss minimum on both the train and test sets.

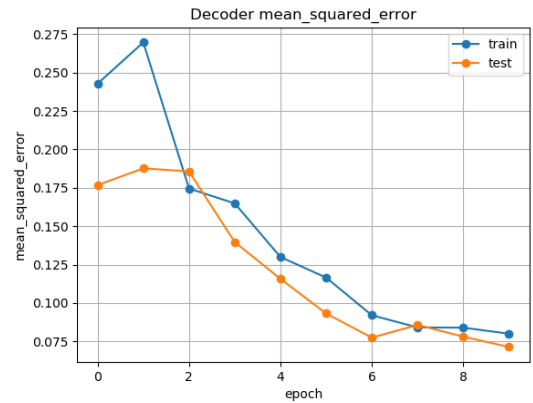

**Figure 2:** Mean square error optimization for the training and test sets for the decoder

Random test images decoded using trained D shows promising results. While the generated images by D are very lossy and noisy they do retain the general structure and major details of the original image. Therefore the theory and principle is validated however any practical use requires further work to improve the quality of decoded generated images. It is to be notes that these results required extensive hyperparameter optimization and some pre and post processing.

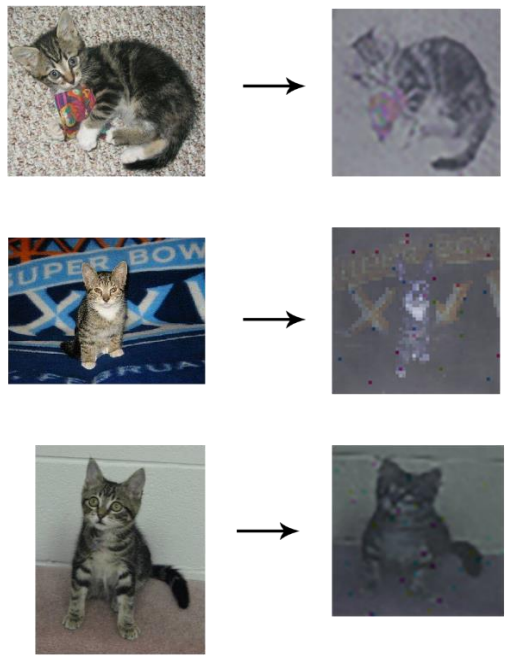

**Figure 3:** Visualization of sample decoded images from the model. Images on the left are original images which were encrypted by the Encoder into the encoded vectors. Images on the right are the corresponding generated images by the Decoder from the encoded vectors. Examples shown here are manually selected as some of the better ones on random test images.

## 7. Advantages & Disadvantages

This new cryptography framework comes with advantages and disadvantages relative to conventional cryptography techniques as well as deep learning techniques. The major advantage is primarily computational. The encoder is able to run on most low power portable devices as stated earlier and the encodings are practically impossible to decrypt without the trained decoder. Another advantage is that the images are encoded into floating point vectors of same size which adds additional security and needs lesser storage.

The major disadvantage is that while the process itself is secure in the sense that the encodings can't be decrypted without the trained decoder, a person can train another decoder to decrypt the encodings if a large number of encodings are available to him. In other words if a device with the encoder and consequently the image encodings generated are available, then another decoder can be trained with a new dataset generated by that encoder. Therefore, while the data (encodings) is safe the encoder itself must be protected from going into the wrong hands. Another disadvantage is the information loss as the decryption process in lossy and the decoded image is of much lower quality as compared to the original image.

## 8. Conclusions and future work

The framework proposed is explained mathematically and tested experimentally with the results shown. This paper has demonstrated the ability of deep learning networks to be able to decrypt encodings of data encrypted from nonlinear functions, suggesting that further research in this area can be useful. This framework can be further extended by:

1. Increasing Decoder complexity: The decoder is relatively simple as compared to current standards. This is due to the focus of this paper is to establish and verify the principles and not on achieving the best quality. Another limitation was the available computation power. The decoder can be made deeper by adding more hidden layers on varying types. CNNs can be explored for both the encoder as well as decoder [23], [24]. Sequential models like RNN, LSTM [25] as well as ensemble networks can increase the quality of decoded images.
2. Increasing dataset size and feature selection: Increasing the dataset size i.e. the image encoding pairs to train the decoder can reduce the *mse* further for better results. Introducing feature selection techniques can also help in reducing the size of encoded vectors and make them more representative of the images themselves [26].
3. Introducing cipher encryption in encoder: Introducing traditional key based encryption techniques in the encoding process along with the neural network can eliminate the stated major disadvantage of the framework while also inhering their own disadvantages.